\documentclass[11pt]{article}
\usepackage{acl2015}
\usepackage{mathptmx}
\usepackage[scaled=.90]{helvet}
\usepackage{courier}
\usepackage{multirow}
\usepackage{amsmath}
\usepackage{booktabs}
\usepackage{url}
\usepackage{microtype}
\usepackage{hyperref}
\usepackage{latexsym}
\newcommand\blfootnote[1]{%
  \begingroup
  \renewcommand\thefootnote{}\footnote{#1}%
  \addtocounter{footnote}{-1}%
  \endgroup
}
\usepackage{xspace}
\newcommand{\eg}{\textit{e.\,g.}\xspace}
\newcommand{\ie}{\textit{i.\,e.}\xspace}
\newcommand{\F}{$\text{F}_1$\xspace}
\usepackage{graphicx}
\makeatletter
\def\@biblabel#1{}
\makeatother

\hypersetup{%
pdftitle={Opinion Mining in Online Reviews About Distance Education Programs},
pdfauthor={Janik Jaskolski, Fabian Siegberg, Thomas Tibroni, Philipp Cimiano, Roman Klinger},
 pdfkeywords={opinion mining, sentiment analysis, distance education, nlp,
   machine learning, natural language processing, text classification,
   social media, digital humanities},
hidelinks
 }

\title{Opinion Mining in Online Reviews About Distance Education Programs}

\author{%
Janik Jaskolski$^{1,2}$, Fabian Siegberg$^3$,\\\bf Thomas Tibroni$^3$, Philipp Cimiano$^{1,2}$, Roman Klinger$^{1,2,4}$\\
\begin{minipage}{0.3\linewidth}\centering
$^1$CITEC\\
Bielefeld University\\
33615 Bielefeld, Germany
\end{minipage}
\begin{minipage}{0.3\linewidth}\centering
$^2$Semalytix\\
Stennerstra\ss{}e 27\\
33613 Bielefeld, Germany\\
\end{minipage}\\[8mm]
\begin{minipage}{0.3\linewidth}\centering
$^3$Online Akademie\\
  Zollstockg\"urtel 63\\
  50969 Cologne, Germany\\
\end{minipage}
\begin{minipage}{0.3\linewidth}\centering
  $^4$IMS\\ 
  Uni Stuttgart\\
  70569 Stuttgart, Germany\\
\end{minipage}\\
\tt jaskolski@semalytix.de,\\\tt
\{f.siegberg,t.tibroni\}@online-akademie.com,\\\tt
cimiano@cit-ec.uni-bielefeld.de, klinger@ims.uni-stuttgart.de
}

\begin{document}
\maketitle
\blfootnote{\hspace{-5.8mm}This work has been performed while the first and last
  authors were at Bielefeld University.}
\begin{abstract}
  The popularity of distance education programs is increasing at a fast
  pace. En par with this development, online communication in fora,
  social media and reviewing platforms between students is increasing
  as well. Exploiting this information to support fellow students or
  institutions requires to extract the relevant opinions in order to
  automatically generate reports providing an overview of pros and
  cons of different distance education programs.  We report on an
  experiment involving distance education experts with the goal to
  develop a dataset of reviews annotated with relevant categories and
  aspects in each category discussed in the specific review together
  with an indication of the sentiment.
  Based on this experiment, we present an approach to extract general
  categories and specific aspects under discussion in a review
  together with their sentiment. We frame this task as a multi-label
  hierarchical text classification problem and empirically investigate
  the performance of different classification architectures to couple
  the prediction of a category with the prediction of particular
  aspects in this category. We evaluate different architectures and
  show that a hierarchical approach leads to superior results in
  comparison to a flat model which makes decisions independently.
\end{abstract}

\section{Introduction}
\label{sec:introduction}

Online and distance education has contributed strongly to the
accessibility to higher education. Next to relatively new players on
the market which focus on distance education like Coursera, edX, and
Udacity, traditional campus universities offer online education
platforms as well.  In Germany, in addition to the largest institution
for distance education (Fern\-universit\"at Hagen, with 88000
registered students \cite{Dahlmann2013}), several private institutions
offer courses, for instance, the Apollon University of Applied Science
and Healthcare or the Euro-FH.

The popularity of online courses is demonstrated through annual
enrollment rates which continue to exceed the growth rates of
traditional higher education \cite{allen2010learning,allen2011going}.
Students face an overwhelming challenge to select a program fitting
their interests due to the high amount of offers available.  Distance
education courses have high drop-out rates that are regularly credited
to poor matches of students to courses
\cite{brinton2013learning}. Finding a program that fits ones needs,
schedule, learning pace, or financial possibilities is therefore of
crucial importance.

An appealing option to find information about a program in addition to
official material is relying on the perspective of other fellow
students that, in contrast to the distance education providers themselves, have no
incentives per se in promoting a particular study program.  In fact, the
information exchanged about online education programs on the Web (\eg
in fora, social media, blogs) has been increasing recently.  In
reaction to this trend, several providers of recommendation services
are turning towards hosting platforms in which students can review
their distance education programs and the corresponding institutions
offering these programs. One example for such a recommendation service
in Germany is the \emph{Online Akademie}
(\url{http://www.fernstudiumcheck.de}, OAK), which maintains
information about 4908 courses from 477 institutes (as of
May $4^{\text{th}}$, 2016).  Similarly, U.S. News and World Reports
(\url{http://www.usnews.com/education/online-education}) generate
ranked lists for different fields of research. The study portal for
distance learning (\url{http://www.distancelearningportal.com/}) is a
worldwide search engine for classes and programs and provides
structured information. Springtest (\url{http://www.springest.de}) is
a platform for individual education that compares 37870
courses (as of May $4^{\text{th}}$, 2016).

This exchange of information between students is hard to be assessed
at large scale, either by the students themselves or by institutions
who could benefit from this feedback to improve their services.  One
major challenge is that the information exchange is mainly in natural
language. Given the fast pace at which the amount of reviews increases
(for instance 4785 new reviews in 2013), a comprehensive, manual
analysis does not scale to the rate at which reviews are produced. An
automated support for analyzing textual reviews automatically is needed.

Towards developing a methodology to analyze such reviews and make
their content accessible, we provide the following contributions in
this paper and highlight how they address the practical use case: We
describe the \emph{methodology} used for the \emph{creation of a
  corpus of user generated reviews for distance education}. We discuss
the agreement on the task between different annotators, and provide a
qualitative analysis.  Further, we present an approach that
automatically extracts the categories and nested aspects within these
categories under discussion in a review together with the sentiment
expressed towards these.  We frame the task as a \emph{multi-label
  hierarchical classification task} in which relevant aspects, grouped
into categories, are given. The task is to predict both the relevant
categories as well as the specific aspects under discussion in a
review. This is therefore a \emph{closed-domain} approach, similarly
to, \eg, the restaurant data set \cite{Ganu2009}. We propose and
investigate architectures that take into account the hierarchical
structure of the categories and experimentally investigate the impact of
classification architectures that couple the prediction of categories
and specific aspects that enable the classifiers to exchange
information and thus to tie their decisions together. We show that,
while a flat model reaches reasonable precision of 79\,\% on the
prediction of aspects (at a recall of 16\,\% and \F measure of
26\,\%), a hierarchical model in which information from category
prediction is propagated to the aspect (``subcategory'') prediction
increases recall to 44\,\% at a precision of 71\,\% and thus an \F
measure of 54\,\%.

\begin{figure*}
  \centering
  \fbox{\includegraphics[width=0.7\textwidth]{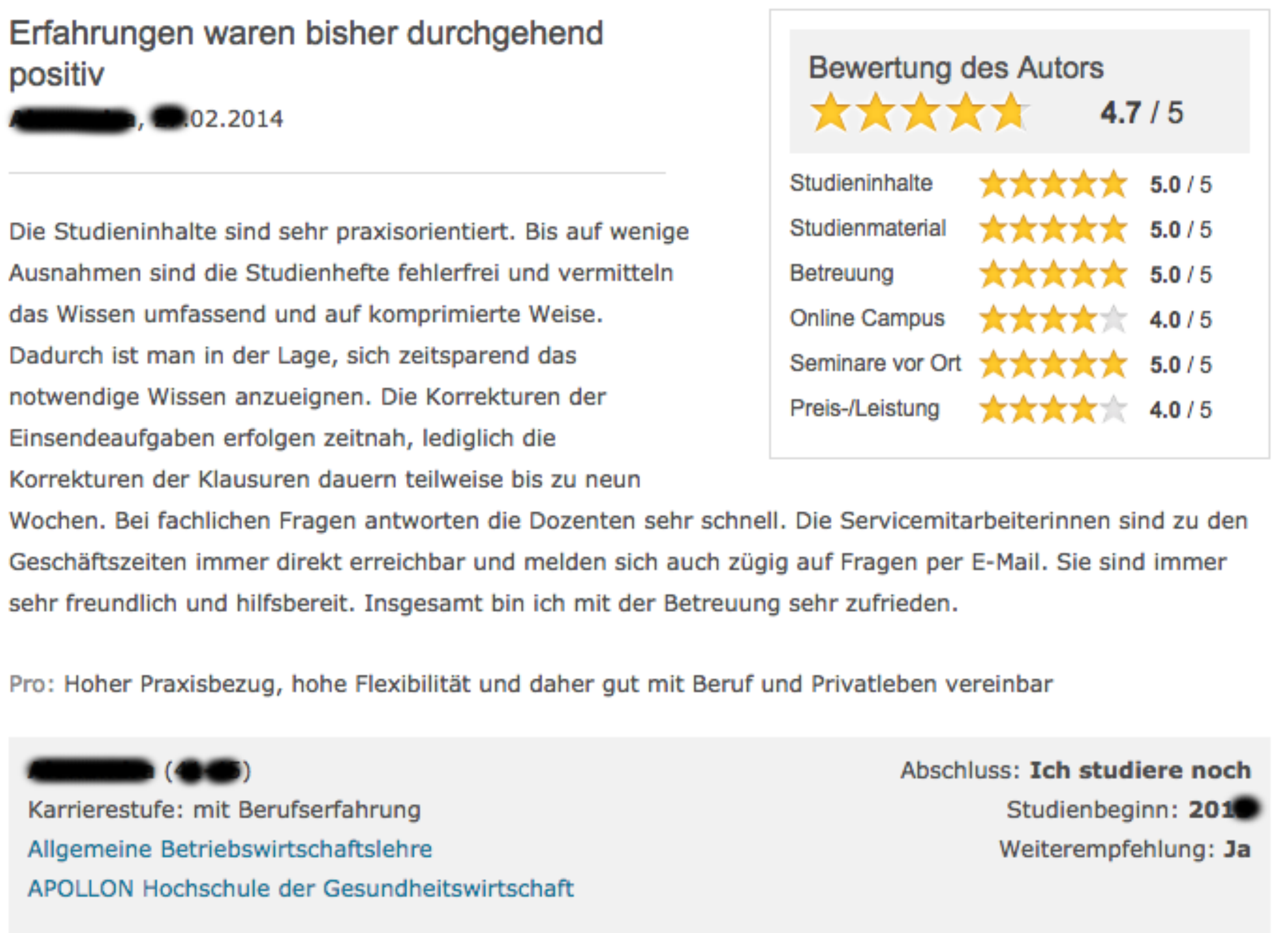}}
  \caption{Example review from the website
    \protect\url{fernstudiumcheck.de/} (potentially identfying
    information has been blacked).}
  \label{fig:ex}
\end{figure*}

\section{Corpus of Distance Education User Reviews}
\label{l:corpus}
The data source taken into account are the free text reviews from the
website \url{http://fernstudiumcheck.de/}.  We randomly sample 394
reviews from the overall set of 9142 reviews, as of Oct.\
$10^{\text{th}}$, 2014. Each sentence is annotated with the aspects
discussed and the opinion expressed by the author towards each
aspect. An example review is shown in Figure~\ref{fig:ex}.

The annotation of the corpus with aspects was performed by three
annotators.  Each sentence is assigned a list of pairs of aspect and
the polarity of this aspect. More formally, each sentence $s_i$ is
associated with a set of tuples $(a^i_j,p^i_j)$.

The polarity $p^i_j$ ranges from $-9$ to $+9$, with $-9$ representing a
\emph{``very negative opinion''} (for instance \emph{``\ldots the worst
experience ever\ldots''}) and $+9$ representing a \emph{``very positive
  opinion''} with respect to the aspect in question (for instance
\emph{``\ldots an outstanding and great\ldots''}). The value $0$ corresponds to
a neutral opinion (\emph{``it is ok''}). The value $99$ has been used in
annotation for a mixed polarity (for instance \emph{``\ldots is good but I
hate it\ldots''}).

We take into account 32 aspects as values for the variables
$a^i_j$. These aspects have been found to cover all relevant
information by preliminary annotation experiments. Each aspect belongs
to one of the eight categories \emph{study contents}, \emph{study
  materials}, \emph{support and organization}, \emph{didactics},
\emph{online campus}, \emph{tuition}, \emph{attendance seminar},
\emph{personal}.  The full list of aspects and categories is shown in
Table~\ref{t:aspectdistribution}.
\begin{table}[t]
\renewcommand{\arraystretch}{0.92}
\setlength{\tabcolsep}{6.5pt}
\centering\footnotesize
\begin{tabular}{lrrr}
\toprule
 \textbf{Class Label} & \textbf{Occ} & \textbf{Pos} & \textbf{Neg} \\
\cmidrule(r){1-1}\cmidrule(rl){2-2}\cmidrule(lr){3-3}\cmidrule(rl){4-4}
\textbf{Study Contents} & \textbf{397} & \textbf{272} & \textbf{138}\\
Average Demand 	& 60 	&      	23	& 	25	\\
Up-To-Date 			& 58 	&       35	&	20	\\
Practical Relevance & 50 	&  	36	&	8	\\
Quality of Contents 	& 229 	&  	143	&	66	\\
Exams 			& 77 	& 	35	&	19	\\
\cmidrule(r){1-1}\cmidrule(rl){2-2}\cmidrule(lr){3-3}\cmidrule(rl){4-4}
\textbf{Study Materials} & \textbf{101} & \textbf{75} & \textbf{20}\\
Production Quality	& 7 		&      	5	& 	2	\\
Accessibility 		& 22 	&       20	&	2	\\
Extent of Materials	& 28 	&  	21	&	3	\\
Exercise Materials 	& 44 	&  	29	&	13	\\
\cmidrule(r){1-1}\cmidrule(rl){2-2}\cmidrule(lr){3-3}\cmidrule(rl){4-4}
\textbf{Support and Orga.} & \textbf{749} & \textbf{594} & \textbf{125}\\
Supervision 			& 487 	&      	409	& 	61	\\
Revision Time 		& 89 	&       78	&	10	\\
Organization 		& 173 	&  	107	&	54	\\
\cmidrule(r){1-1}\cmidrule(rl){2-2}\cmidrule(lr){3-3}\cmidrule(rl){4-4}
\textbf{Didactics} & \textbf{527} & \textbf{384} & \textbf{110}\\
Teaching Competence 	& 124 	&      	101	& 	13	\\
Didactics of Materials 	& 308 	&       220	&	72	\\
Justified Grading	 	& 20		&  	15	&	3	\\
Revision Quality		& 75 	&  	48	&	22	\\
\cmidrule(r){1-1}\cmidrule(rl){2-2}\cmidrule(lr){3-3}\cmidrule(rl){4-4}
\textbf{Online-Campus} & \textbf{196} & \textbf{129} & \textbf{47}\\
Usefulness 		& 119 	&      	84	& 	28	\\
Activity 			& 35 	&       29	&	6	\\
User-Friendliness	& 20 	&  	8	&	11	\\
Features			& 22 	&  	18	&	2	\\
\cmidrule(r){1-1}\cmidrule(rl){2-2}\cmidrule(lr){3-3}\cmidrule(rl){4-4}
\textbf{Tuition} & \textbf{90} & \textbf{55} & \textbf{20}\\
Basic Tuition 		& 78 	&      	48	& 	17	\\
Additional Charges 	& 10 	&       7	&	3	\\
Scholarships	 	& 2 		&  	0	&	0	\\
\cmidrule(r){1-1}\cmidrule(rl){2-2}\cmidrule(lr){3-3}\cmidrule(rl){4-4}
\textbf{Attendance Seminar} & \textbf{127} & \textbf{93} & \textbf{22}\\
Seminar Contents 	& 84 	&      	59	& 	14	\\
Management	 	& 18 	&      12	&	5	\\
Locations	 		& 9 		&  	7	&	2	\\
Communications	& 16 	&  	15	&	1	\\
\cmidrule(r){1-1}\cmidrule(rl){2-2}\cmidrule(lr){3-3}\cmidrule(rl){4-4}
\textbf{Personal} & \textbf{572} & \textbf{472} & \textbf{74}\\
Flexibility			 	& 103 	&      	96	& 	2	\\
Recommendation	 	& 77 	&       71	&	4	\\
Personal Benefit 		& 74 	&  	61	&	6	\\
Overall Satisfaction		& 236 	&  	215	&	17	\\
Learning Effort			& 82 	&  	29	&	45	\\
\cmidrule(r){1-1}\cmidrule(rl){2-2}\cmidrule(lr){3-3}\cmidrule(rl){4-4}
\textbf{Other} & \textbf{345} & \textbf{0} & \textbf{0}      \\
No Label			 	& 345 	&      0	& 	0	\\
\bottomrule
\end{tabular}
\caption{Statistics of the annotations of the corpus of 394 reviews.}
\label{t:aspectdistribution}
\end{table}
%
A summary of the statistical properties of the corpus is presented in
Section \ref{sec:corpusstatistics}.

\section{Models for Automatically Estimating Aspect-Polarity Tuples}
\label{l:models}
We phrase the task of assigning aspects and polarities to sentences as
a text classification problem in which dependencies between different
classes are reflected in and captured by the model.  We first describe
the various features used to take a decision with regards to which category or aspect is
under discussion in a particular review. Then, we describe the different
classification models we investigate.

\subsection{Features for Aspect Categorization}
In order to classify each sentence as to whether it describes a given
aspect or not, our model relies on tf$\cdot$idf scores for each term
\cite{Manning2008}.  In addition, we compute the occurrences of word
bigrams and trigrams of all sequences of a length of two and three of
directly succeeding words. As domain-specific features, the occurrence
of terms in manually compiled dictionaries containing specific words
frequently associated with a specific aspect, category, or polarity
orientation are taken into account.

\subsection{Features for Polarity Detection}
As cues for polarity tokens and phrases, the English dictionary by
\newcite{Hu2004} was automatically translated via Google Translate
(\url{https://translate.google.com}) to German. The feature computed
on the basis of this list is the number of polarity words in a
sentence. The occurrence of diminishing and intensifying nouns is
estimated using manually compiled domain-specific word lists.  The
GermanPolarityClues dictionary \cite{WALTINGER10.91} is used to
determine sentiment priors for individual words and summed up to
create an aggregated sentiment score for the entire sentence for
positive, negative, and neutral sentiment.  Bigrams are used as
features as described for aspect detection. For unigrams, negations
are taken into account by building features from the cross-product of
terms like \emph{not} and \emph{n't}, upon first occurrence, with all
succeeding words in the same sentence.

\subsection{Model Structures}

In our classification approach, we use one binary classifier for each
aspect.  Each of these classifiers is trained to predict whether the
corresponding aspect is mentioned in the sentence the classifier is
applied to.  In addition, the aspects are grouped into categories. 

Assigning a broader category is often easier for a human than
assigning a specific category. Further, knowing which category is
discussed in a given sentence might simplify the task of predicting
which aspects are specifically under discussion. Therefore, we extend
the classification-based approach to the category level and train
classifiers that predict whether any aspect of a given category is
discussed in the respective sentence.

The two classifications can be coupled, so that the classifier
performing a decision on the specific aspect under discussion has
access to the decision of the classifier predicting which more general
category is discussed in the sentence.  Coupling these decisions can
be implemented via different classifier architectures that we describe
below.  Similarly, the binary classifiers predicting whether a
sentence is neutral, positive or negative can be coupled.

\begin{figure*}[t]
  \centering
  \includegraphics[scale=0.7,page=2]{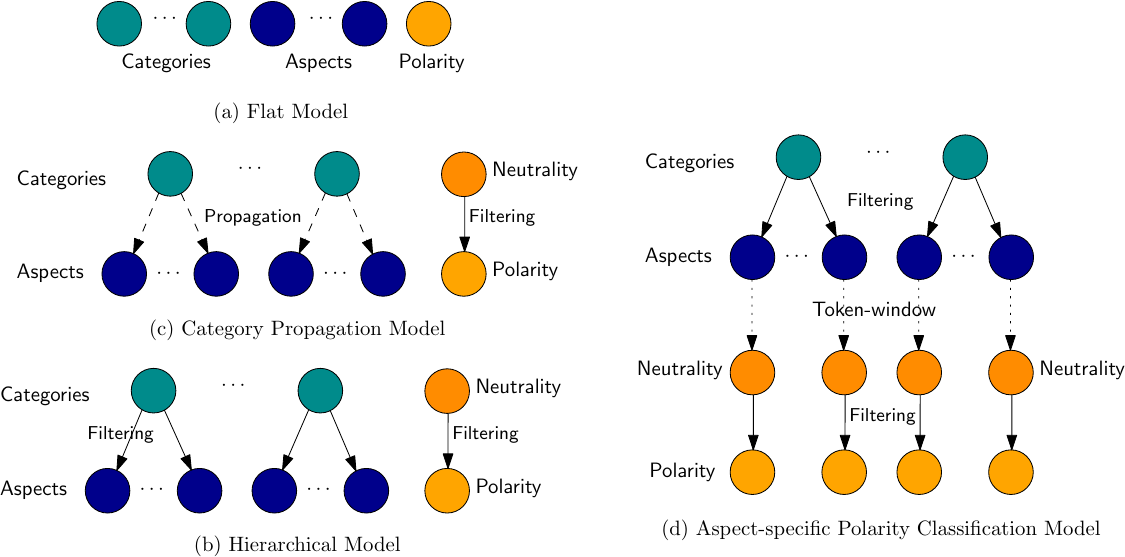}
  \caption{Visualization of the different model structures. The \emph{Flat
    Model} (a) does not take into account dependencies between aspects
    or categories. In the \emph{Hierarchical Model} (b), instances are only
    classified by aspect classifiers if the corresponding category
    classifier made a positive decision. In the \emph{Category Propagation
    Model} (c), the output of the category classifiers are used as
    features in the aspect classifiers. The polarity classifiers are
    used in the same manner as in the Hierarchical Model. In the
    \emph{Aspect-Specific Polarity Classification Model} (d), the context
    defined by an aspect prediction is taken into account for polarity
    prediction.}
  \label{figureTwo}
\end{figure*}

In the \emph{Flat Model} (see Figure~\ref{figureTwo}), each
category and each aspect are predicted independently from each other,
so that there is no interaction between these
classifications. Similarly, the polarity of a sentence is predicted
independently of the prediction of the aspect or category classifiers.
This model is considered a baseline structure.

In the \emph{Hierarchical Model}, all
classifiers for the categories are applied first.  An aspect
classifier is then only applied if the respective category classifier
has predicted that the sentence in question is about the category the
aspect belongs to.  Besides this hierarchical dependence, there is no
direct information flow between the different classifiers.
As an example, a prediction that a sentence belongs to the category
\emph{Tuition} would prompt the classifiers \emph{Basic Tuition},
\emph{Scholarships}, and \emph{Accessory Charges} to be applied. If
the initial prediction of the \emph{Tuition} classifier is negative,
none of the above mentioned aspect classifiers would be applied. The polarity detection is independent of the aspect detection
in this model, but only applied if a previous model decides for a
sentence to be non-neutral.
The intuition of this model is that the classifiers on the higher
level (categories, neutral/non-neutral) can take into account
properties shared between sentences of classes of the
lower level (aspects, polarities). On the other hand, classifiers on
the lower level can focus on more specific properties and are
therefore only trained on sentences of the associated categories.

The \emph{Category Propagation Model}
uses the same structure as the \emph{Hierarchical Model}. However,
instead of only applying classifiers on the lower level to sentences
which have been `let through' by the upper level, all classifiers are
always applied. The information from upper levels is propagated to the
lower level by additional features. 

The intuition is that the hierarchical model might be too
strict. Errors on an upper level would propagate to lower levels. In
the propagation model, the predicted categories are used as
features in the aspect classifiers. Therefore, these models can `vote
against' the decision of the previous level.
However, the task is more challenging for the aspect classifiers, as
they take into account all aspects, not only the ones from the same
category. The polarity classification remains deterministic and is
incorporated in the same manner as before.

In the \emph{Aspect-Specific Polarity Classification Model}, polarity classifiers make the
decision based on the textual context of the mention of the respective
aspects. Therefore, in this model, different aspects can be assigned
different polarities in contrast to the other models. For this purpose, for each aspect the terms with
the highest mutual information are assumed to express
the respective aspect. The polarity classifier is then only applied on
a window of $n$ tokens left and right of the informative tokens for an
aspect. For instance, the sentence \emph{``The lecturer was very
  entertaining but the course work was dreadful''} would be assigned two aspects with opposite polarities.

\section{Results}
\label{sec:evaluation}
In this section, we describe the properties and distributions of aspects in a manually annotated corpus. This information is valuable on
the one hand as it presents results which can be assumed to hold for
the whole corpus of all reviews as the subset has been sampled
uniformly. On the other hand, it presents results on the task of automatically
detecting such aspects in more detail.

\subsection{Corpus Statistics and Observations}
\label{sec:corpusstatistics}

The corpus has been annotated by three annotators following guidelines
that have been agreed upon and incrementally refined along four
annotation rounds.  The final inter-annotator agreement reached during
these rounds has been substantial with a $\kappa=0.75$
\cite{Cohen1960}). In the final corpus, every third review has been
annotated by two annotators for quality assurance ($\kappa=0.77$).

The corpus consists of 394 reviews with 2481 sentences. On average, a
review consists of $6.3$ sentences. Each sentence was annotated on average
with 1.25 aspect-polarity pairs, leading to 3103 annotations
altogether.

Table \ref{t:aspectdistribution} summarizes the overall frequencies of
all aspects, categories and polarities. The category with the most
frequent mentions is \emph{Support and Organization} (749 mentions),
closely followed by \emph{Personal} aspects (572) and \emph{Didactics}
(527). These categories can therefore be considered to be of high
importance. Surprisingly, the \emph{Tuition} was by far less often
discussed (90). Ordering the categories by polarity (\ie, by
``positive''$-$``negative'' annotations), this ranked list is not
changed, positive mentions are by far more frequent than negative
aspect mentions. The category that is most frequently discussed in a negative way is
\emph{Study Content}, though this is only the fourth most
frequently mentioned category. When normalizing the absolute numbers
of mentions by the overall numbers of mentions of a category, the
\emph{Study Content} remains on the top of the list of negatively mentioned
categories (34\,\% of all mentions are negative), closely
followed by the \emph{Quality of Content} (29\,\%).  The most frequent categories discussed positively are \emph{Personal} (70\,\%) and
\emph{Support and Organization} (62\,\%).

Going to the more fine-grained level, the \emph{Flexibility},
\emph{Communications}, \emph{Recommendation}, \emph{Overall
  Satisfaction}, \emph{Accessibility}, \emph{Revision Time},
\emph{Personal Benefit}, \emph{Features } (of online campus),
\emph{Support}, \emph{Teaching Competence} are the top 10 aspects
with respect to the positive vs. negative ratio.
(with percentages of positive mentions ranging from 94\,\% to
81\,\%). On the other side, the categories most frequently discussed in a negative manner are the \emph{Learning Effort}, the
\emph{User-Friendliness}, \emph{Average Demand}, \emph{Scholarships},
\emph{Exams}, \emph{Up-To-Dateness}, \emph{Organization},
\emph{Revision Quality}, \emph{Exercise Materials} and
\emph{Management} (with percentages of positive mentions ranging from
35\,\% to 66\,\%).

Notably, there are only three aspects which are more often mentioned
negatively than positively, namely the \emph{Average Demand}, the
\emph{User-Friendliness}, and the \emph{Learning Effort}.

\begin{figure}[!t]
  \centering
  \includegraphics[scale=0.77]{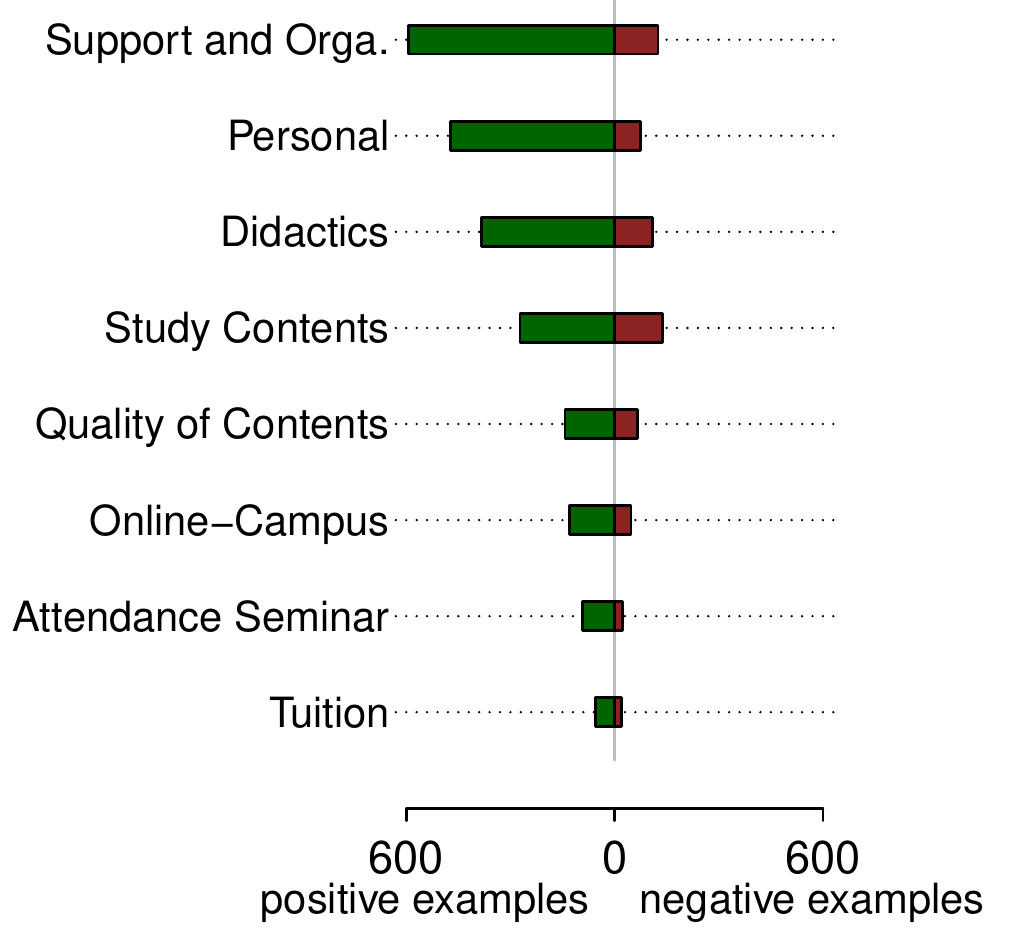}
  \caption{Distribution of positive and negative annotations for individual categories, ranked in a descending order by ratio of positive vs. negative examples.}
  \label{figureThree}
\end{figure}

It is important to mention that this data set is heavily unbalanced,
both in the distribution of aspects as well as in the distribution of positive
and negative mentions, as depicted in Figure~\ref{figureThree}.

\subsection{Model Evaluation}
In the following, we analyze the performance of the different
classification model structures with respect to  the task of predicting the
aspects and the polarity of sentences automatically.

\paragraph{Experimental Setting}
The evaluation is performed on an independent test set of 104
reviews. The remaining 290 reviews are used for training the
models. Note that this split has been the same for the whole
engineering phase. The 104 test instances have therefore never been
used to improve the performance of the automated system.  As
performance measures, precision, recall, and \F measure are used.

\emph{Precision} measures the correctness of a prediction of a system:
how many of the aspects detected by the system are actually correct?
\emph{Recall} expresses how many of the aspects that a sentence
actually mentions are also found by the automatic classification
system.  The F$_1$-Measure is the harmonic mean of these
two measures.  Formally, the precision of a system on some data set
with respect to one specific aspect is defined as
$\mathrm{prec}=\frac{\mathrm{TP}}{\mathrm{TP}+\mathrm{FP}}$, the
recall is
$\mathrm{rec}=\frac{\mathrm{TP}}{\mathrm{TP}+\mathrm{FN}}$. The \F
measure is the balanced harmonic mean of these two values:
$F_1=2\frac{\mathrm{prec}\cdot\mathrm{rec}}{\mathrm{prec}+\mathrm{rec}}$. TP,
FP, and FN denote the number of true positives, false positives, and
false negatives, respectively.

The macro-averaged precision, recall, and \F measure values are
calculated from the respective values for the different aspects. The
micro-averaged value is calculated based on the sum of TP, FP, FN for
all aspects as the basis for the performance measures. Therefore, the
micro-average takes into account the different distributions of the
aspects, while the macro-average does not.

\paragraph{Aspects and Categories}
The results for the aspect and category detection are summarized in
Table~\ref{t:modelcomparison}. The micro-averaged values tend to be
higher than the macro-averaged scores due to the unbalancedness of the
aspect frequencies; this means that the more frequent an aspect is,
the more accurately it can be detected. However, the differences
between different model structures are similar, such that we focus on
the micro-averaged scores in the discussion.
\newcommand{\one}[1]{\multicolumn{1}{c}{#1}}
\begin{table}
\renewcommand{\arraystretch}{1.2}
\setlength{\tabcolsep}{1.5pt}
\centering\footnotesize
\begin{tabular}{lccccccccc}
\toprule
&&& \multicolumn{3}{c}{Macro (\%)} && \multicolumn{3}{c}{Micro (\%)} \\
\cmidrule(r){3-5}\cmidrule{6-8}
Model & Type && P & R & F && P & R & F \\
\cmidrule(r){1-1}\cmidrule(lr){2-2}\cmidrule(r){4-6}\cmidrule{8-10}
\multirow{2}{*}{\rotatebox{0}{Flat}}
&\one{Categories} && 70&35&46&&72&44&55\\
&\one{Aspects}   && 79&16&26&&72&27&39\\
\cmidrule(r){1-1}\cmidrule(lr){2-2}\cmidrule(r){4-6}\cmidrule{8-10}
\multirow{3}{*}{\rotatebox{0}{Hier.}}
&\one{Categories}   && 70&35& 46&&72&44&55\\
&\one{Categ. (Infer.)} && 71&31&43&&72&38&50\\
&\one{Aspects}   && 71&{44}&{54}&&61&{70}&{65}\\
\cmidrule(r){1-1}\cmidrule(lr){2-2}\cmidrule(r){4-6}\cmidrule{8-10}
\multirow{3}{*}{\rotatebox{0}{Prop.}}
&\one{Categories} && 70&35&46&&72&44&55\\
&\one{Categ. (Infer.)} && 73&29&41&&75&36&49\\
&\one{Aspects}   && 74&{17}&{28}&&65&{30}&{41}\\
\bottomrule
\end{tabular}
\caption{Results for different model structures. The results for the
  categories are as predicted for the flat model (``Flat''). In the
  hierarchical model (``Hier.'') and the category propagation model 
  (``Prop.''), the category labels as inferred from the aspect 
  prediction are provided in addition.}
\label{t:modelcomparison}
\end{table}

The results for the categories are the same for all model types as
this decision is always made independently of other classifiers, i.e.,
there is no information flow from the aspect classifiers to the
category classifiers, but only the other way round.  The rows labeled
with ``Categories (Infer.)'' are the performance measures when
inferring the category labels from the aspects. For the hierarchical
model and the category propagation model, a category is regarded to be
relevant for a sentence if at least one classifier for an aspect within this category votes for the sentence. Clearly, these inferred values are inferior to
the specialized category classifiers. However, when outputting
categoric results to a user, contradictions in the results should be
avoided and therefore the inferred category labels would be
preferred. This occurs particularly in the category propagation model,
proving that the capability to overwrite a decision by the category classifier
leads to an improved precision (0.75 vs.\ 0.72).

Altogether, for the category detection, precision is higher than
recall. This is the case for the prediction of aspects as well, but to
a smaller extent.  For the aspect recognition, the flat model has the
lowest performance with 0.39 \F followed by the category propagation
model with 0.41 \F. The hierarchical model improves this result by
0.24 percentage points to 0.65 \F.  However, the best precision is
reached with the flat model. The label propagation model represents a
tradeoff between precision and recall with an \F-Measure of 0.41.  It
can thus be concluded that the use of categorical information for the
aspect recognition enables the aspect classifier to focus on the distinction between
aspects of the same category and thus increases recall.

Analyzing results for the respective classifiers, it is notable that
specifically the performance for aspects with a small number of
training instances is limited: The best results are achieved for
\emph{usefulness, supervision, basic tuition, revision time}, and
\emph{recommendation} (with an average \F of 0.81) with 850 instances. The
worst results are achieved for \emph{accessibility, additional charges,
communication in attendance seminars, features of the online campus,
locations of the attendance seminars} with 79 instances altogether.

\paragraph{Polarities}
The results for the aspect-independent polarity detection (as in the
first three model structures) is shown in
Table~\ref{t:agnosticvsspecific}. The detection whether one sentence
contains a positive or negative statement is close to perfect. The low
number of neutral sentences is therefore more difficult to be
detected. However, in this task the detection of positive and negative
statements is of far greater importance. The detection of positive
sentences is satisfactory, with an \F of 84\,\%. The recognition of
negative sentences turns out to be more challenging, with 61\,\% \F.

\begin{table}
\centering\footnotesize
\begin{tabular}{lcccccccc}
\toprule
&&\multicolumn{3}{c}{Agnostic (\%)}&&\multicolumn{3}{c}{Specific (\%)}\\
\cmidrule(r){3-5}\cmidrule{7-9}
&& P & R & F && P & R & F \\ 
\cmidrule(r){3-5}\cmidrule{7-9}
Positive && 78&92&84&&76&91&83 \\ 
Negative && 63&60&61&&57&40&47   \\ 
Neutral && 75&18&29&&86&10&18  \\ 
Polar && 88&99&93&&93&100&96  \\ 
\bottomrule
\end{tabular}
\caption{Results for polarity and neutrality detection, both 
  aspect-agnostic and aspect-specific.}
\label{t:agnosticvsspecific}
\end{table}

In addition, the results show that there is added benefit for
distinguishing polar from neutral sentiment for specific aspects
compared to computing the polarity at the level of the whole sentence.
However, overall, this is a rare phenomenon; only 119 sentences out of
2481 mention more than one aspect. Spanned across the entire problem,
this rarity leads to difficulties in learning from the increased
granularity.  This is expressed in experiments with different window
sizes. For aspect-specific detection of all polarities, maximal
performance is reached when using the entire sentence for
classification.

The aspect-agnostic polarity classification approach delivers correct
results in 99.95\% of the cases.

\section{Discussion}
The analysis of the manual annotation of the corpus clearly proves
that students have a tendency to discuss positive aspects rather than
negative aspects of their distance education programs. This is a bit
surprising as one might expect a bias to report negative rather than
positive experiences.  However, one observes a clear tendency towards
discussing positive rather than negative aspects. The relatively high fees of distant education programs might bias people towards a more positive assessment.

In general, the discussions typically revolve around a few categories
specific for distance education programs that are generally evaluated
positively.  The most frequent categories discussed negatively 
comprise of learning procedures, that is \emph{materials, demand,
  effort, and quality}. This might suggest that distance education
programs are specifically good in their domain but must improve on
providing an as solid learning environment as attendance-focused
institutions.

The corpus discussed in this work has been used to automatically learn
models of different structure. The best performing model indicates
that the process of detecting a category first followed by the
corresponding aspects is beneficial. The results are satisfactory for
frequent aspects.

Our assessment of different model structures discovered that sentences
of the same category share properties, which explains the increased
recall for the hierarchical model.  The best precision is however
achieved by the flat model which makes decisions for all aspects
independently of each other.

The main use case of our method is to support the processing of
reviews in real-time so that feedback about distance education
programs can be taken into account in shorter time to support decision
making by prospective students. For instance, interested students can
select one of the aspects to rank programs and institutions
based on that selection. In addition, evidence for specific strengths
or weaknesses can be easily retrieved in the form of example reviews
or sentences.

In addition to a benefit for potential students, such an approach can
help institutions to identify their specific weaknesses. For instance, in order
detect which of their programs has most problems regarding
management, material quality or other aspects.

\section{Conclusion and Future Work}

The number of reviews on online education programs is increasing
steadily. The reviews from peers contain valuable information that can
help students to find the distance education programs that fits their
needs best. Given the unstructured nature of such reviews, a
systematic and large scale analysis is challenging.

The analysis of the content of such reviews to make their content
accessible in a structured manner is a real-world use case for text
mining, sentiment analysis and natural language processing.  We have
proposed a methodology for the automatic and systematic analysis of
these reviews. On the one hand, we have proposed a methodology for the
manual annotation of reviews in the form of annotation guidelines
comprising a taxonomy of aspects that are relevant to distance
education programs. On the other hand, we have developed an automatic
approach that can automatically identify the aspects under discussion
in a given review, and make the sentiment towards these aspects
explicit. This automatic approach exploits classifiers trained using
machine learning techniques to identify whether a given aspect is
discussed in the review under consideration or not. In particular,
different model structures for these classifiers have been empirically
evaluated. We have shown that the results provided by the classifiers
are satisfactory and support the automation of the task. The
performance of the polarity detection is satisfying. The recall for
aspects of lower frequency in our corpus has space for
improvement. However, it can be expected that increasing the size of
the corpus with a focus on the infrequent aspects will allow for for an
overall improvement of the classification.

Future work includes the enlargement of the corpus and an analysis of
the impact of the performance of the classifiers on actual ranking of
service providers with respect to specific aspects. The actual
acceptance of users of websites supporting the decision which program
suits them well will be evaluated.

In addition, the findings should be compared to structured information
available for the programs and institutions. One example is the
question: Is there a correlation between satisfaction and happiness
with the tuition fees and the actual amount of money the program
costs?

Furthermore, the current method of propagating information through the
hierarchy does not include propagation bottom-up. We will compare the
impact of a feed-forward neural network-like setting to address this
potential limitation in the current approach.

\newpage

\bibliographystyle{acl}
\bibliography{literature}

\begin{thebibliography}{}

\bibitem[\protect\citename{Allen and Seaman}2010]{allen2010learning}
I.~Elaine Allen and Jeff Seaman.
\newblock 2010.
\newblock {\em Learning on Demand: Online Education in the United States,
  2009}.
\newblock Babson Survey Research Group.

\bibitem[\protect\citename{Allen and Seaman}2011]{allen2011going}
I.~Elaine Allen and Jeff Seaman.
\newblock 2011.
\newblock {\em Going the Distance: Online Education in the United States,
  2010}.
\newblock Babson Survey Research Group.

\bibitem[\protect\citename{Brinton \bgroup et al.\egroup
  }2014]{brinton2013learning}
C.~Brinton, M.~Chiang, S.~Jain, H.~Lam, Z.~Liu, and F.~Wong.
\newblock 2014.
\newblock Learning about social learning in moocs: From statistical analysis to
  generative model.
\newblock {\em Learning Technologies IEEE Transactions on Learning
  Technologies}, PP(99):1--1.

\bibitem[\protect\citename{Cohen}1960]{Cohen1960}
Jakob Cohen.
\newblock 1960.
\newblock {A Coefficient of Agreement for Nominal Scales}.
\newblock {\em Educational and Psychological Measurement}, 20(1):37.

\bibitem[\protect\citename{Dahlmann}2013]{Dahlmann2013}
Wolfgang Dahlmann.
\newblock 2013.
\newblock {Deutschlands gr\"o\ss{}te Uni}.
\newblock {\em Unispiegel}.
\newblock Online: \url{http://spon.de/ad6HQ}.

\bibitem[\protect\citename{Ganu \bgroup et al.\egroup }2009]{Ganu2009}
Gayatree Ganu, Noemie Elhadad, and Am\'elie Marian.
\newblock 2009.
\newblock Beyond the stars: Improving rating predictions using review text
  content.
\newblock In {\em International Workshop on the Web and Databases}.

\bibitem[\protect\citename{Hu and Liu}2004]{Hu2004}
Minqing Hu and Bing Liu.
\newblock 2004.
\newblock Mining and summarizing customer reviews.
\newblock In {\em Proceedings of the Tenth ACM SIGKDD International Conference
  on Knowledge Discovery and Data Mining}, KDD '04, pages 168--177, New York,
  NY, USA. ACM.

\bibitem[\protect\citename{Manning \bgroup et al.\egroup }2008]{Manning2008}
Christopher~D. Manning, Prabhakar Raghavan, and Hinrich Sch\"{u}tze.
\newblock 2008.
\newblock {\em Introduction to Information Retrieval}.
\newblock Cambridge University Press, New York, NY, USA.

\bibitem[\protect\citename{Waltinger}2010]{WALTINGER10.91}
Ulli Waltinger.
\newblock 2010.
\newblock Germanpolarityclues: A lexical resource for german sentiment
  analysis.
\newblock In Nicoletta Calzolari~(Conference Chair), Khalid Choukri, Bente
  Maegaard, Joseph Mariani, Jan Odijk, Stelios Piperidis, Mike Rosner, and
  Daniel Tapias, editors, {\em {Proceedings of the Seventh International
  Conference on Language Resources and Evaluation (LREC'10)}}, pages
  1638--1642, Valletta, Malta, may. European Language Resources Association
  (ELRA).

\end{thebibliography}

\end{document}